\documentclass[11pt]{article}

\usepackage[final]{acl}

\usepackage{times}
\usepackage{latexsym}

\usepackage[T1]{fontenc}

\usepackage[utf8]{inputenc}

\usepackage{microtype}

\usepackage{inconsolata}

\usepackage{graphicx}

\usepackage{amssymb} 
\usepackage{multirow} 

\usepackage[most]{tcolorbox}  

\tcbset{
  aibox/.style={
    enhanced,
    colback=gray!5,
    colframe=black!15,
    fonttitle=\bfseries,
    coltitle=black,
    boxrule=0.5pt,
    arc=2mm,
    outer arc=2mm,
    top=1mm,
    bottom=1mm,
    left=2mm,
    right=2mm,
    title={#1}
  }
}

\title{S2S-Arena: Evaluating Paralinguistic Instruction Following
in Speech-to-Speech Models}

\author{
  \textbf{Feng Jiang\textsuperscript{1}},
  \textbf{Zhiyu Lin\textsuperscript{2}},
  \textbf{Yiyang Liu\textsuperscript{3,1}},
  \textbf{Liumeng Xue\textsuperscript{3}},
  \\
  \textbf{Fan Bu\textsuperscript{2,4}},
  \textbf{Yuhao Du\textsuperscript{2,4}},
  \textbf{Xiangying Chen\textsuperscript{5}},
  \textbf{Benyou Wang\textsuperscript{2,4}}\thanks{Benyou Wang is the corresponding author.},
  \textbf{Haizhou Li\textsuperscript{2,4,6}},
\\
  \textsuperscript{1} Artificial Intelligence Research Institute, Shenzhen University of Advanced Technology \\
  \textsuperscript{2} The Chinese University of Hong Kong, Shenzhen \\
  \textsuperscript{3} Nanjing University
  \textsuperscript{4} Shenzhen Loop Area Institute \\
  \textsuperscript{5} CentraleSupélec, Université Paris-Saclay,
  \textsuperscript{6} National University of Singapore \\
  \small{
    \textbf{Correspondence:} \href{mailto:email@domain}{jiangfeng@suat-sz.edu.cn}, \href{mailto:email@domain}{wangbenyou@cuhk.edu.cn}
  }
}

\begin{document}
\maketitle
\begin{abstract}
Recent advances in large language models (LLMs) have fundamentally reshaped speech-to-speech (S2S) systems, enabling increasingly natural spoken interaction. 
However, existing benchmarks still rely heavily on text-based evaluation and largely ignore paralinguistic cues such as prosody, emotion, and speaker traits, which are central to expressive and human-like communication. 
We introduce S2S-Arena, a speech-native benchmark for evaluating instruction-following S2S models with explicit assessment of both semantic understanding and paralinguistic expression. S2S-Arena features a four-level interaction protocol that systematically probes models under increasing paralinguistic complexity, a two-stage data construction pipeline that produces 1,243 speech samples spanning 100+ real-world tasks, and an arena-style evaluation framework that enables reference-free, pairwise comparison directly in the speech modality. 
Benchmarking 10 state-of-the-art S2S systems over 1,000+ comparisons reveals substantial performance gaps (especially under complex paralinguistic demands) between current academic and industrial systems. Our analysis further identifies key design factors governing expressive instruction following, providing actionable insights for building more natural, robust, and human-aligned speech agents.

\end{abstract}

\section{Introduction}

Voice-based human-computer interaction offers one of the most natural and intuitive modalities for communication~\cite{card1983psychology, allen2001toward}. In speech-to-speech (S2S) systems, models are expected not only to understand spoken input~\citep{chu2023qwen, chu2024qwen2, tangsalmonn, GAMA, wavllm}, but also to generate appropriate spoken output and complete tasks~\cite{wang2024maskgct, chen2024f5, liao2024fish}.

\begin{figure}[t]
    \centering
    \includegraphics[width=1\linewidth]{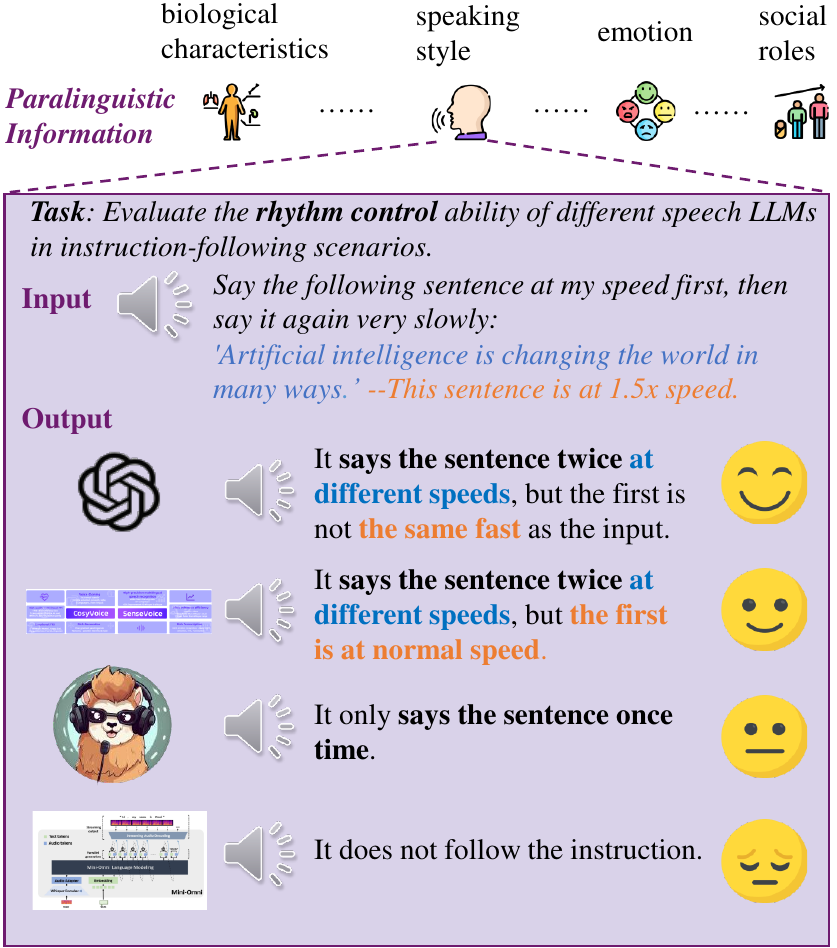}
    \caption{An example of evaluating rhythm-aware instruction following in both speech input and output.}
    \label{fig:example}
\end{figure}

Recent advances in large language models (LLMs)~\citep{dubey2024llama} have driven the development of expressive S2S systems~\citep{speechGPT, speechteam2024funaudiollm, fang2024llama, xie2024mini}, shifting attention from pure semantics to paralinguistic features, such as prosody, emotion, speaking style, and speaker identity~\cite{schuller2010interspeech, nose2007style, batliner2011automatic, ipgrave2009language}. As illustrated in Figure~\ref{fig:example}, these signals are essential for human-like interaction. Although semantic accuracy ensures task completion, this paralinguistic information conveys empathy, intent, and social appropriateness, which are more crucial for real-world applications such as education, emotional support, and medical consultation. 

As shown in Table~\ref{table:benchmarks}, the existing evaluations for S2S systems~\citep{dynamic_superb, audiobench, saibench} are gradually focusing on paralinguistic information, but with a greater emphasis on speech understanding rather than generation, emphasizing the instruction following ability of paralinguistic information. There are also some works that attempt to evaluate aspects such as emotion understanding in the chat scenario~\cite{sdeval} or attempt to output in sound modality~\citep{chen2024voicebenchbenchmarkingllmbasedvoice, airbench}, but regardless of the method, they assess speech outputs only via transcripts, missing the fidelity of paralinguistic expression~\citep{ji2024wavchat}.

To establish a diagnostic benchmark with quantified uncertainty and bias analysis, providing actionable guidance for the design of future speech-to-speech systems, we introduce the \textsc{S2S-Arena} benchmark.
We first design a four-level evaluation framework with increasing difficulty, covering four representative application scenarios (education, entertainment, social interaction, and medical consultation) and 19 task categories. This framework not only encompasses speech understanding tasks considered in prior benchmarks but also extends evaluation to paralinguistic expression in speech generation, providing a more holistic assessment of S2S capabilities.

We further develop a two-stage data construction pipeline to ensure both scale and quality. Starting from a manually curated seed set, we expand the dataset through speech-native self-instruction, generating additional audio samples in the speech modality. This process yields a final benchmark of 1,243 audio-based queries specifically designed to probe paralinguistic understanding and generation.

Finally, we conduct over 1,000 large-scale pairwise comparisons of 10 popular speech-to-speech (S2S) models under a scalable arena-style evaluation framework. Our results reveal a substantial performance gap between industrial and academic systems. We further provide an in-depth analysis from both task-domain and task-difficulty perspectives, uncovering how key factors (training data, speech encoders, backbone language models, and speech decoders) contribute to S2S performance. These findings offer valuable insights that lay a solid foundation for future research in this area.

Our contributions are fourfold:

We introduce the first speech-native arena-style benchmark for S2S models that explicitly evaluates both semantic correctness and paralinguistic expressiveness.

We propose a four-level interaction protocol that formalizes progressive paralinguistic reasoning and generation in speech interaction.

We construct a large-scale benchmark (1,243 samples, 100+ tasks) via a two-stage pipeline combining expert curation and speech-native self-instruction.

Through 1,000+ speech-native pairwise comparisons over 10 state-of-the-art models, we reveal systematic capability gaps and identify key architectural factors governing expressive instruction following.

\section{Related Work}
\label{sec:rw}
\subsection{LLM-based S2S Models}
\label{sec:s2s-models-parainfo}

Recent speech-to-speech (S2S) systems have evolved from classical cascaded pipelines toward increasingly integrated end-to-end architectures.
Early systems typically follow an ASR$\rightarrow$LLM$\rightarrow$TTS paradigm~\cite{speechteam2024funaudiollm}, where linguistic content and paralinguistic traits are handled by separate modules. More recent work consolidates speech understanding and generation into a unified model that can be trained jointly.
Despite architectural diversity, modern LLM-based S2S systems consistently revolve around three core components:
a speech encoder, an LLM backbone, and a speech decoder. More details are shown in Appendix~\ref{app:s2s-models}.

\begin{table*}[!ht]
    \centering
    \resizebox{\linewidth}{!}{
    \begin{tabular}{llcccccccc}
        \hline
        \textbf{Benchmarks for Speech Models} & \textbf{Task Types} & \multicolumn{2}{c}{\textbf{Understanding}} & \multicolumn{2}{c}{\textbf{Generation}} & \multicolumn{2}{c}{\textbf{Evaluation}} \\ 
        & & \textbf{Sem.} & \textbf{Par.} & \textbf{Sem.} & \textbf{Par.} & \textbf{Modality} & \textbf{Evaluator} \\
        \hline
        Dynamic-SUPERB~\citep{dynamic_superb} & Foundation  & \checkmark  & \checkmark  & \checkmark*  &   & - *  & LLM \\ 
        SGAI~\cite{saibench} & Foundation & \checkmark  & \checkmark  &    &    & Text  & LLM \\ 
        AudioBench~\citep{audiobench} & Foundation & \checkmark  & \checkmark  &    &    & Text  & LLM \\ 
        MMAU~\citep{sakshi2024mmau} & Foundation & \checkmark  & \checkmark  &    &    & Text  & LLM \\ 
        AV-Odyssey Bench~\citep{gong2024av} & Foundation & \checkmark  & \checkmark  &    &    & Text  & LLM \\ 
        Vstyle~\cite{zhan2025vstyle} & Foundation & \checkmark  &  &   \checkmark  &   Style  & Speech  & LALM \\ 
        \hline
        SD-Eval~\citep{sdeval} & Chat & \checkmark  & \checkmark  &    &    & Text  & LLM \\ 
        Voxdialogue~\cite{cheng2025voxdialogue} & Chat & \checkmark  & \checkmark  &    &    & Text  & LLM \\ 
        VoiceBench~\citep{chen2024voicebenchbenchmarkingllmbasedvoice} & Chat & \checkmark  &    & \checkmark  &    & Text  & LLM \\ 
        \hline
        AIR-Bench~\citep{airbench} & Mixed & \checkmark  & \checkmark  & \checkmark  &    & Text  & LLM \\ 
        Multivox~\cite{selvakumar2025multivox} & Mixed & \checkmark  & \checkmark  & \checkmark  &     & Text  & LLM \\ 
        S2S-Arena (Ours) & Mixed & \checkmark  & \checkmark  & \checkmark  & \checkmark & Speech & Human/S2S model \\ 
        \hline
\end{tabular}
    }
    \caption{Comparison of Benchmarks for Speech2Speech Models. The star* means that the evaluation modality of the Dynamic-Superb is decided by the tested task. Sem. means the semantics of speech, and Par. means the paralinguistic information of speech, such as biological characteristics, speaking style (such as pitch, tone, speed), emotion, and social roles (such as background and age). }
    \label{table:benchmarks}
\end{table*}

\paragraph{Speech Encoders.}
From the encoder perspective, current models fall into two main categories.
Speech-token-based systems (e.g., SpeechGPT~\cite{speechGPT}, GLM-4-Voice~\cite{zeng2024glm}, Kimi-Audio~\cite{ding2025kimi}, Baichuan-Omni-1.5~\cite{li2025baichuan}) discretize speech into tokens using vector quantization (VQ or RVQ), often combined with Whisper-large encoders.
In contrast, speech-embedding-based systems (e.g., Mini-Omni~\cite{xie2024mini}, LLaMA-Omni~\cite{fang2024llama}, Qwen2.5-Omni~\cite{xu2025qwen2}) preserve continuous acoustic embeddings by attaching lightweight adaptors to Whisper encoders.

\paragraph{LLM Backbones.}
On the language modeling side, most systems are built upon popular large pretrained LLMs such as Qwen~\cite{qwen2025qwen25technicalreport}, LLaMA~\cite{dubey2024llama}, or GLM~\cite{glm2024chatglm}. Backbone choice strongly influences instruction following, reasoning, and controllability of paralinguistic behaviors.

\paragraph{Speech Decoders.}
For speech generation, recent systems (e.g., GLM-4-Voice, Kimi-Audio, Baichuan-Omni-1.5, Qwen2.5-Omni) increasingly adopt flow-matching based generative models combined with neural vocoders such as HiFi-GAN~\cite{kong2020hifi} or BigVGAN~\cite{leebigvgan}.
Other designs integrate non-autoregressive (NAR), autoregressive (AR), or codec-based decoders (e.g., Freeze-Omni). Decoder architecture critically affects speech naturalness, prosody control, and expressive range.

\subsection{Benchmarks for Paralinguistic Evaluation}

With the rapid progress of S2S models, evaluation benchmarks have gradually evolved from only speech understanding or speech generation~\cite{zhang2025speechjudge} to S2S instruction following, as shown in Table~\ref{table:benchmarks}.

Benchmarks such as Dynamic-SUPERB~\citep{dynamic_superb}, SGAI~\cite{saibench}, MMAU~\citep{sakshi2024mmau}, AudioBench~\citep{audiobench}, and AV-Odyssey~\citep{gong2024av} design specific tasks (e.g., emotion recognition, speaker identification, background inference) that force models to attend to paralinguistic signals in speech inputs. Vstyle~\cite{zhan2025vstyle} attempts to focus on the style of speech output under the constraint instruction input with Audio-Language Model as Judge.

More recent benchmarks move toward more realistic interactive settings.
Chat-style evaluations such as SD-Eval~\citep{sdeval}, Voxdialogue~\cite{cheng2025voxdialogue} and VoiceBench~\citep{chen2024voicebenchbenchmarkingllmbasedvoice} adopt dialogue-based scenarios in which models must reason over everyday speech containing rich paralinguistic cues.
However, both continue to rely on text-based evaluation and do not assess whether generated speech faithfully preserves paralinguistic attributes.

A small number of benchmarks further extend evaluation from understanding to generation. For example, AIR-Bench~\citep{airbench} and Multivox~\cite{selvakumar2025multivox} incorporate prosody-aware instructions and some speech generation tasks, yet still evaluate outputs primarily via transcripts.

Overall, existing benchmarks largely treat paralinguistic information as an auxiliary input feature for understanding, while the quality and faithfulness of paralinguistic expression in generated speech, a core requirement of real-world S2S systems, remain severely underexplored.

\section{S2S-Arena}
\label{sec3}

As discussed in Section~\ref{sec:rw}, recent advances in speech-to-speech (S2S) modeling have enabled increasingly natural spoken interaction.
However, existing benchmarks remain fundamentally limited in two critical aspects:
(1) they emphasize speech \emph{understanding} while largely neglecting the evaluation of expressive \emph{speech generation}; and
(2) their evaluation pipelines operate primarily in the \emph{text modality}, inevitably discarding rich paralinguistic information such as prosody, emotion, speaking style, and speaker traits.

\begin{figure*}[t]
    \centering
    \includegraphics[width=1\linewidth]{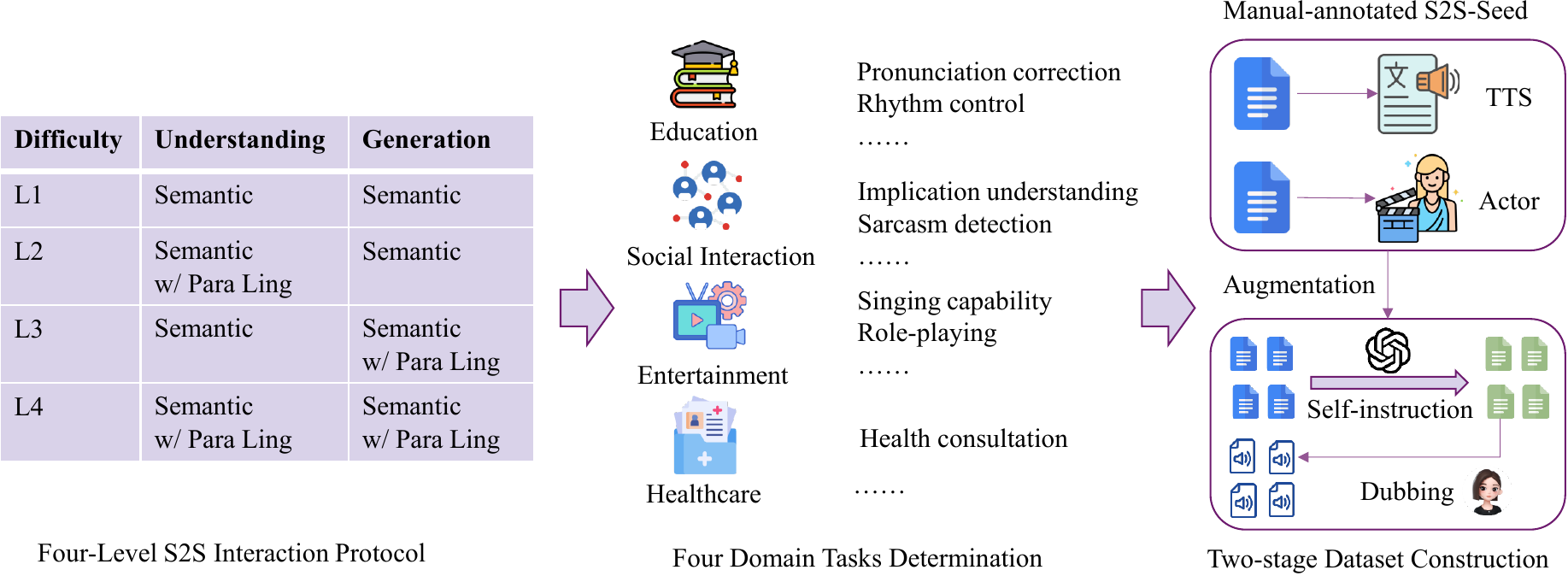}
    \caption{Overview of the S2S-Arena construction pipeline.}
    \label{fig:model}
\end{figure*}

To address these limitations, we introduce \textbf{S2S-Arena}, a comprehensive benchmark for evaluating S2S systems \emph{directly in the speech modality}, with explicit consideration of both semantic correctness and paralinguistic expressiveness in realistic human-like interaction.

As shown in Figure~\ref{fig:model}, the construction of \textsc{S2S-Arena} follows a unified design philosophy that integrates three tightly coupled components:
(i) a four-level hierarchical interaction framework that progressively evaluates S2S models from basic semantic instruction following to full paralinguistic interaction with increasing difficulty;
(ii) a systematic task design over four application domains and 19 representative tasks that captures diverse real-world S2S scenarios; and
(iii) a two-stage dataset construction pipeline that combines carefully curated human seed data with large-scale automatic self-instruction to form a diverse, scalable, and realistic benchmark dataset.

\subsection{Four-Level S2S Interaction Protocol}

To systematically characterize the expressive capabilities of S2S models, we design a hierarchical interaction protocol that decomposes speech interaction into four progressively challenging levels.
Unlike prior benchmarks that focus primarily on semantic instruction following, our protocol explicitly models the joint evolution of \emph{semantic understanding} and \emph{paralinguistic competence}, reflecting how humans naturally communicate in spoken dialogue.

As illustrated in Figure~\ref{fig:model}, the four levels are defined as follows:

\textbf{L1: Instruction-Only.}
This level evaluates pure semantic understanding and task execution, without considering any paralinguistic factors.
For example, the user asks: ``I have a headache. What could be the cause?''
A correct response provides medically plausible explanations, independent of tone, emotion, or speaking style.

\textbf{L2: Paralinguistic Perception.}
The model must \emph{perceive} paralinguistic cues from the input speech and adapt its semantic response accordingly, while its own output remains neutral.
For example, a child asks: ``If it rains tomorrow, how should I plan my day?''
The model should infer the speaker’s age from vocal cues and generate age-appropriate advice.

\textbf{L3: Paralinguistic Expression.}
Here the input speech is semantically neutral, but the instruction explicitly requires the model to \emph{express} specific paralinguistic attributes in the output.
For example, the user asks: ``Recite a tongue twister at three different speeds: fast, medium, and slow.''
The model must control acoustic properties such as tempo in the generated speech.

\textbf{L4: Full Paralinguistic Interaction.}
This most realistic setting requires the model to jointly \emph{perceive} paralinguistic cues from the input and \emph{generate} correspondingly expressive output.
For example, the user asks in a cheerful tone: ``Tell him in Chinese that Mike will arrive tomorrow.''
The model is expected to produce a Chinese translation with matched cheerful prosody.

\subsection{Four-Domain Tasks Design}

Following prior benchmarks~\cite{sdeval, airbench}, we identify four representative application domains:
\textit{Education}, \textit{Social Interaction}, \textit{Entertainment}, and \textit{Medical Consultation}. Then, we conduct a task elicitation study with 19 researchers, each proposing five typical S2S tasks per domain.

From these candidates, we curate 19 representative tasks that explicitly require both semantic reasoning and paralinguistic competence, including pronunciation correction~\cite{speechocean762}, emphasis control~\cite{du2024cosyvoice}, emotional expression~\cite{emotionalTTS}, and singing~\cite{liu2022diffsinger}.
The full task list is provided in Appendix (Table~\ref{tab:sample}).

\subsection{Two-Stage Dataset Construction}

To faithfully instantiate the proposed interaction protocol at scale, we design a two-stage dataset construction pipeline (Figure~\ref{fig:model}) combining human crafting and auto generation that balances annotation quality, coverage diversity, and scalability.

\subsubsection{Manual Seed Data Collection}
We first manually collect few samples in each task under 4-level difficulty constraints to construct high-quality seed dataset. For each task, we design at least 10 scripts with dubbing or manual recording. Specifically, neutral samples without considering paralinguistic information are synthesized using Doubao-TTS, while samples involving strong paralinguistic cues (e.g., emotion, speaker identity, speaking style) are produced via human recording or selected from high-quality corpora such as the Ryerson Emotional Speech Dataset~\cite{livingstone2018ryerson}.
To enhance realism, we further introduce eight categories of background noise (e.g., airport, street, café).

All audio samples undergo strict human quality control by four native Mandarin annotators (2 male, 2 female), each with IELTS speaking scores above 6.5.
Samples with perceptual artifacts, unclear paralinguistic expression, or semantic mismatch are discarded. After checking, we keep 293 high-quality seed samples spanning 19 tasks.

\subsubsection{Automatic Self-instruction Augmentation}

To scale the benchmark while preserving the interaction structure and paralinguistic richness of the seed data, we employ an iterative few-shot self-instruction strategy~\cite{wang2023self}. We define a JSON-formatted script and use GPT-4o to generate new scripts with a 5-shot prompting strategy. More details are shown in Appendix~\ref{app:audio_script}.

The generated scripts are synthesized into speech using controllable TTS systems (Doubao-TTS, AudioX~\cite{tian2025audiox}, and Parler-TTS~\cite{lyth2024natural})  selected for their strong modeling of prosody, emotion, and acoustic variation.
To maintain natural diversity and robustness, no additional filtering is applied, allowing moderate noise, style variation, and minor imperfections to remain. This procedure generates 50 additional samples per task, yielding 950 automatically constructed examples. 

Furthermore, we conduct an additional manual verification study to assess alignment quality. To ensure that the synthetic samples faithfully reflect both the intended evaluation level (L1–L4) and the designated paralinguistic attributes, we randomly sample 25 instances from each level (100 samples in total). The Difficulty Level consistency is 90\% agreement, and Paralinguistic consistency is 93\% agreement. These findings indicate that the large majority of synthetic samples align well with their intended design, while still retaining moderate natural variability.

\subsection{Data Statistics}

The final S2S-Arena contains 1,243 speech samples~\footnote{We release the source data of the S2S-Arena at \url{https://github.com/FreedomIntelligence/S2S-Arena}.}, with the level-wise distribution summarized in Table~\ref{tab:s2s-level}. 
The newly generated Augment dataset comprises over 100 tasks, with 75.79\% of the samples being English, 24\% being Chinese, and the remaining 0.21\% being Japanese and Hindi. We give a detailed task distribution in the Appendix~\ref{app:dataset}.

\begin{table}[htbp]
\centering
\resizebox{\linewidth}{!}{
\begin{tabular}{l|c|cccc|c}
\hline
\textbf{S2S-Arena} & \textbf{\#Tasks} & \textbf{L1} & \textbf{L2} & \textbf{L3} & \textbf{L4} & \textbf{Total} \\
\hline
Seed     & 19     & 32  & 81  & 80  & 100 & 293 \\
Augment  & 100+   & 49  & 248 & 395 & 258 & 950 \\
\hline
All      & 100+   & 81  & 329 & 475 & 358 & 1243 \\
\hline
\end{tabular}
}
\caption{Level-wise distribution of S2S-Arena samples.}
\label{tab:s2s-level}
\end{table}

\section{Experiments}

\subsection{Arena-style Evaluation in the Speech Modality}

A central design principle of \textsc{S2S-Arena} is to evaluate models \emph{directly in the speech modality}.
Most existing benchmarks convert model outputs into text and rely on text-based LLM judges~\cite{zheng2023judging}, which inevitably discard crucial paralinguistic information, including prosody, emotion, speaking style, and speaker traits~\cite{chen2024humans, ye2024justice, speechGPT}.
To preserve these expressive signals, we adopt an arena-style evaluation framework that performs speech-native, reference-free pairwise comparison.

\paragraph{Model Pairing Strategy.}
Rather than uniformly sampling model pairs, we bias sampling toward pairs with \emph{moderate} rating differences.
For all candidate pairs $(i,j)$, we compute the Elo gap $\Delta_{ij} = |R_i - R_j|$ and assign each pair a sampling weight
\begin{equation}
    w_{ij} = \exp\!\left(-\frac{(\Delta_{ij}-\mu)^2}{2\sigma^2}\right),
\end{equation}
where $\mu$ is set to one third of the maximum observed gap and $\sigma = \max(\mu/2, 1)$ controls the smoothness of the distribution.
Pairs are then sampled proportionally to $w_{ij}$.
This strategy avoids trivial comparisons between nearly identical models and uninformative matches between severely mismatched systems, leading to faster and more stable convergence of Elo scores.

\paragraph{Speech-native Judging.}
Unlike prior work that performs judgment in the text modality, all evaluations in \textsc{S2S-Arena} are conducted directly on speech.
We first validate the reliability of automatic evaluation through a preliminary study on the \textsc{S2S-Arena} Seed dataset, involving 19 human annotators for manual judgment and two strong automatic judges (Gemini~2.5-Pro and Qwen2.5-Omni).

For human evaluation, we design a web-based interface in which annotators compare two speech responses generated by anonymous models and decide which better satisfies the spoken instruction under a given task guideline.
For automatic evaluation, each instance consists of three audio segments: (1) spoken task instructions and (2--3) the two candidate model responses.
These are concatenated into a single audio prompt and evaluated jointly under three criteria: instruction alignment, paralinguistic expressiveness, and output audio quality.
Both human and automatic judges produce a strict win/loss decision without ties.

\begin{table}[ht] \centering \begin{tabular}{lcc} \hline \textbf{Evaluator} & \textbf{Kappa} & \textbf{Agreement} \\ \hline Gemini 2.5-Pro & 0.6553 & 82.87\% \\ Qwen2.5-Omni & 0.4667 & 73.15\% \\ \hline \end{tabular} \caption{Agreement between automatic evaluators and human judgments.} \label{tab:auto_eval_agreement} \end{table}

After our comparison, Gemini~2.5-Pro achieves high consistency with human judgments (Cohen’s $\kappa=0.6553$, agreement $82.87\%$), as shown in Table~\ref{tab:auto_eval_agreement}.
We therefore adopt Gemini~2.5-Pro as the automatic judge for large-scale evaluation on the \textsc{S2S-Arena} Augment dataset.
Additional details of the preliminary study are provided in Appendix~\ref{app:elo_rank}.

\paragraph{Elo Score Updating.}
All models are initialized with rating $R=1000$.
For each comparison, the expected score of model $A$ against $B$ is computed as
\begin{equation}
E_A = \frac{1}{1 + 10^{\frac{R_B - R_A}{400}}},
\end{equation}
and the rating is updated by
\begin{equation}
R_A' = R_A + K (S_A - E_A),
\end{equation}
where $S_A \in \{0,1\}$ denotes the outcome and $K$ is fixed to 32.

\subsection{Overall Elo Ranking and Insights}

We benchmark a total of ten representative speech-to-speech (S2S) systems released between 2023 and 2025, selected to reflect the current landscape. They are six industry models (GPT-4o-realtime\footnote{gpt-4o-realtime-preview-2024-10-01} Doubao\footnote{\url{https://www.volcengine.com/docs/6561/1594356}}, FunAudioLLM~\citep{speechteam2024funaudiollm}, GLM-4-Voice~\cite{zeng2024glm}, Qwen2.5-Omni~\cite{xu2025qwen2}\footnote{We use the 7B open-source version.} and Kimi-Audio~\cite{ding2025kimi}) and four academia models (SpeechGPT~\cite{speechGPT}, Mini-Omni~\citep{xie2024mini}, Mini-Omni2~\cite{xie2024mini2}, and LLaMA-Omni~\cite{fang2024llama}). The experimental settings are shown in Appendix~\ref{app:Experimental Details}.

We conduct 1001 pairwise comparisons between these 10 models on the S2S-Arena Augment dataset, with the overall ranking shown in Table~\ref{tab:s2s_elo_rank}~\footnote{To further address concerns regarding statistical reliability, we conducted 1,000 bootstrap resampling runs over match-level comparisons. The bootstrap Elo means preserve the original ranking order. The Spearman correlation (original vs. bootstrap rankings): 0.94 ± 0.03.}. As in other arena-style benchmarks, S2S-Arena supports continual updates: new models can be added and ranked via incremental comparisons.

\begin{table}[!ht]
\centering
\resizebox{\linewidth}{!}{
\begin{tabular}{l r r r r r}
\hline
\textbf{Model} & \textbf{Elo} & \textbf{Win Rate} & \textbf{W} & \textbf{L} & \textbf{Matches} \\
\hline
Qwen 2.5-Omni      & \textbf{1246.1} & 59.0\% & 134 & 93  & 227 \\
GPT-4o-realtime & 1239.2 & 65.7\% & 140 & 73  & 213 \\
Doubao         & 1231.9 & \textbf{67.9\%} & 133 & 63  & 196 \\
GLM-4-Voice    & 1148.2 & 58.3\% & 119 & 85  & 204 \\
FunAudioLLM    & 1088.3 & 51.0\% & 128 & 123 & 251 \\
Kimi-Audio    & 1056.7 & 49.3\% & 142 & 146 & 288 \\
LLaMA-Omni     &  908.7 & 44.4\% &  68 &  85 & 153 \\
Mini-Omni2     &  727.4 & 33.1\% &  59 & 119 & 178 \\
SpeechGPT      &  677.1 & 27.3\% &  42 & 112 & 154 \\
Mini-Omni      &  676.4 & 26.1\% &  36 & 102 & 138 \\
\hline
\end{tabular}}
\caption{Elo scores, win rates, and match statistics for S2S models. Win Rate = ( of pairwise matches won) / (total matches participated). W/L denotes wins/losses. }
\label{tab:s2s_elo_rank}
\end{table}

First, top-tier S2S systems from industry excel along different axes of interaction quality rather than a single dominant dimension:
Qwen~2.5-Omni achieves the highest overall Elo score (1246.1);
GPT-4o-realtime records the largest number of wins (140);
Doubao exhibits the highest win rate (67.9\%). Below this leading group, GLM-4-Voice, FunAudioLLM, and Kimi-Audio form a tightly clustered middle tier, with Elo scores between 1056 and 1148.
Their comparable performance aligns well with their architectural similarities, while the remaining differences are largely attributable to variations in backbone models, speech encoders, and speech decoders, aspects that we analyze in depth in Section~\ref{sec:analysis}. However, a pronounced performance divide between industrial and academic S2S development. LLaMA-Omni stands out as the closest competitor to industrial models, trailing the leaders by roughly 150 Elo points, whereas other academic systems fall behind by over 300 points.

\begin{figure}[h]
    \centering
    \includegraphics[width=\linewidth]{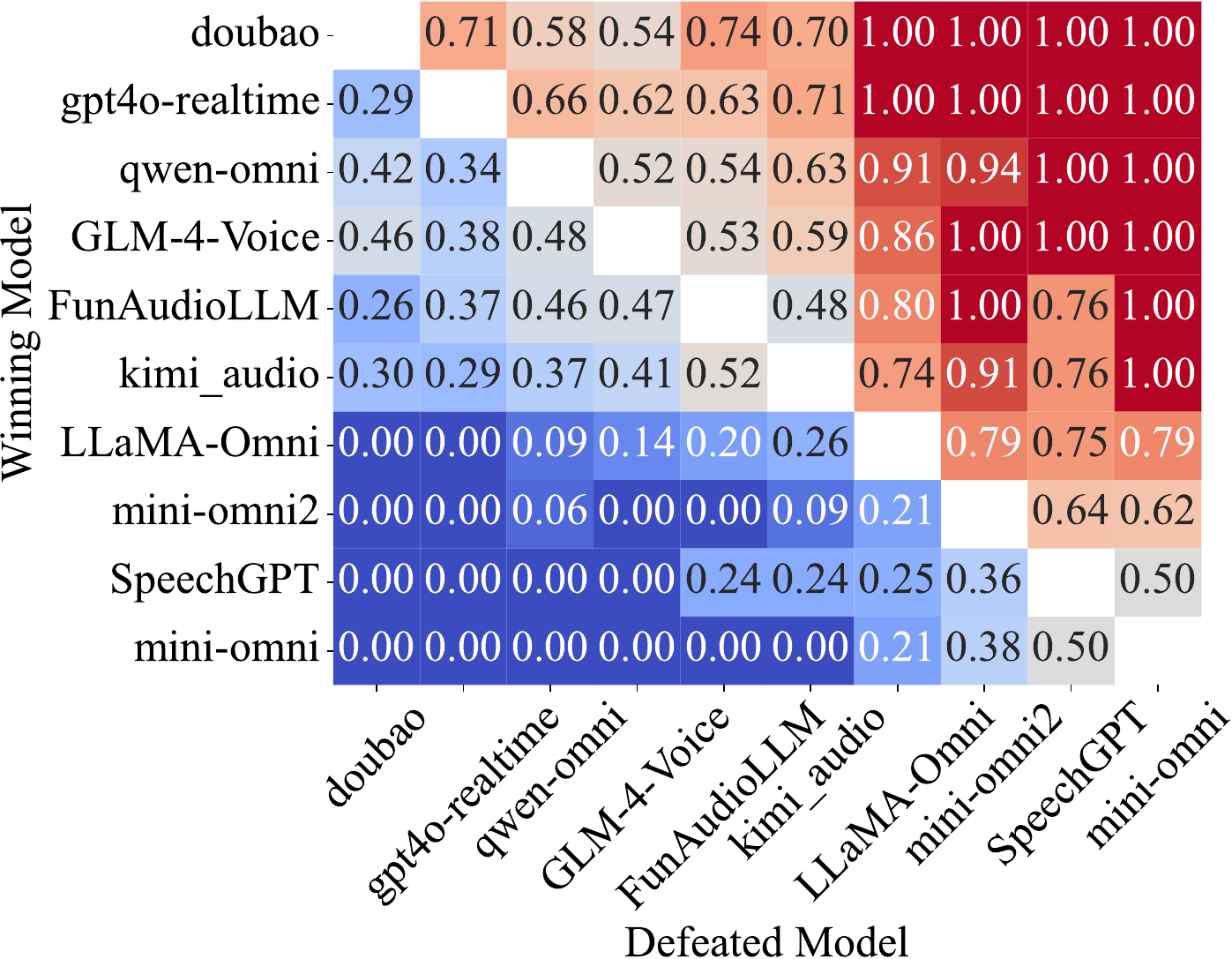}
    \caption{Pairwise win rate matrix among 10 S2S models. Each cell shows the win rate (ranging from 0 to 1) of the row model over the column model.}
    \label{fig:matrix}
\end{figure}

To further examine fine-grained interaction behavior, we visualize head-to-head win rates among all models in Figure~\ref{fig:matrix}.
The heatmap confirms that Doubao consistently outperforms both GPT-4o-realtime and Qwen~2.5-Omni, while the middle tier, GLM-4-Voice, FunAudioLLM, and Kimi-Audio display non-transitive outcomes, each excelling in different pairwise comparisons. The results reveal a clear polarization pattern: dominant models remain dominant, while weaker models differentiate themselves through distinct and specialized strengths.

\subsection{Case Study}
To better understand how models leverage paralinguistic information, we analyze both winning and failing cases across high-performing and underperforming systems. 

We first conduct an in-depth comparison between two high-performing models, Doubao and GPT-4o-realtime. Our analysis reveals that GPT-4o-realtime tends to adopt a more rational and solution-oriented approach when responding to user queries, focusing on providing accurate and informative answers rather than relying on paralinguistic strategies to satisfy users. In contrast, although Doubao exhibits relatively weaker knowledge capabilities, it leans toward using colloquial and expressive paralinguistic cues to enhance user satisfaction. This paralinguistic richness is a likely factor contributing to Doubao's occasional wins over GPT-4o-realtime, despite the latter’s superior knowledge capacity.

In addition, we analyzed cases where relatively strong models, such as Kimi-Audio, underperformed compared to weaker models like SpeechGPT. Interestingly, while Kimi-Audio delivers reasonably good audio quality, its voice often sounds weak or feeble. Although this may enhance human-likeness to some extent, it lacks the clarity and robustness observed in SpeechGPT.

\section{Further Analysis}
\label{sec:analysis}
\subsection{Performance across Task Categories}

To understand how different types of real-world interactions stress distinct S2S capabilities, we analyze model behavior across four domain categories: \textit{Education}, \textit{Entertainment}, \textit{Medical}, and \textit{Social}, as shown in Table~\ref{tab:category_elo}. 

\begin{table}[h]
\centering
\resizebox{\linewidth}{!}{
\begin{tabular}{lrrrrr}
\hline
\textbf{Model} & \textbf{Edu.} & \textbf{Enter.} & \textbf{Medical} & \textbf{Social} & \textbf{Avg.} \\
\hline
GPT-4o-realtime & \textbf{1230.2} & \textbf{1166.8} & \textbf{1124.4} & 1056.6 & \textbf{1144.5} \\
Doubao & 1214.5 & 1144.6 & 1055.7 & 1133.0 & 1136.9 \\
Qwen 2.5-Omni & 1096.7 & 1097.0 & 1056.0 & \textbf{1155.9} & 1101.4 \\
GLM-4-Voice & 1143.2 & 1063.4 & 1096.4 & 1063.8 & 1091.7 \\
FunAudioLLM & 999.3 & 1105.9 & 876.2 & 1123.3 & 1026.2 \\
Kimi-Audio & 1028.1 & 1100.1 & 998.1 & 955.8 & 1020.5 \\
LLaMA-Omni & 922.3 & 1004.6 & 948.3 & 913.6 & 947.2 \\
Mini-Omni2 & 805.2 & 799.3 & 989.7 & 823.9 & 854.5 \\
Mini-Omni & 784.4 & 773.8 & 933.8 & 876.7 & 842.2 \\
SpeechGPT & 776.1 & 744.5 & 921.4 & 897.4 & 834.9 \\
\hline
\end{tabular}
}
\caption{Elo scores across task categories with macro average.}
\label{tab:category_elo}
\end{table}

\begin{table*}[!ht]
\centering
\resizebox{\linewidth}{!}{
\begin{tabular}{llrrrrrlll}
\hline
\textbf{Source} & \textbf{Model} & \textbf{L1} & \textbf{L2} & \textbf{L3} & \textbf{L4} & \textbf{Avg} & \textbf{Backbone} & \textbf{Encoder} & \textbf{Decoder} \\
\hline
\multirow{6}{*}{\textbf{Industry}} 
& GPT-4o-realtime & \textbf{1064.4} & \textbf{1199.2} & \textbf{1241.7} & 1071.3 & \textbf{1144.2}  & - & - & -\\ 
& Doubao & 1029.5 & 1163.7 & 1148.2 & \textbf{1205.8} & 1136.8 & - & - & - \\
& Qwen 2.5-Omni & 1072.2 & 1109.1 & 1136.2 & 1123.0 & 1110.1 & Qwen-2.5 7B & Whisper-large & FM + BigVGAN \\
& GLM-4-Voice & 1053.1 & 1086.3 & 1082.6 & 1093.8 & 1079.0 & GLM-4 9B  & Whisper-large +VQ & FM + HiFiGAN \\
& FunAudioLLM & 957.5 & 1013.1 & 1056.0 & 1087.1 & 1028.4  & Qwen-2 72B & SenseVoice & FM + HiFiGAN \\
& Kimi-Audio & 980.6 & 949.7 & 1130.1 & 1050.7 & 1027.8  & Qwen-2.5 7B & Whisper-large+VQ & FM + BigVGAN \\
\hline
\multirow{4}{*}{\textbf{Academia}} 
& LLaMA-Omni & 977.7 & 965.2 & 920.2 & 942.4 & 951.4 & LLaMA-3.1 8B & Whisper-large & HiFi-GAN  \\
& Mini-Omni & 985.8 & 803.0 & 769.8 & 835.7 & 848.6  & Qwen-2.5 0.5B  & Whisper-small &  SNAC Audio Decoder\\
& Mini-Omni2 & 955.3 & 831.3 & 793.7 & 796.2 & 844.1 & Qwen-2.5 0.5B & Whisper-small & SNAC Audio Decoder \\
& SpeechGPT & 923.9 & 879.4 & 721.5 & 794.0 & 829.7 & LLaMA 7B & mHubert & HiFi-GAN\\
\hline
\end{tabular}
}
\caption{Elo scores across difficulty levels with macro average.}
\label{tab:level_elo}
\end{table*}

\textbf{Finding 1: Knowledge-driven domains emphasize semantic grounding and response reliability.}  
Education and Medical tasks require accurate reasoning, structured explanation, and risk-aware response generation.
Models achieve their strongest performance in these domains when they exhibit robust semantic grounding and factual consistency.
For example, GPT-4o-realtime reaches 1230.2 in Education and 1124.4 in Medical, while Doubao achieves 1214.5 and 1055.7, respectively.

\textbf{Finding 2: Expressive domains reward paralinguistic flexibility and conversational naturalness.}  
Entertainment and Social interactions impose weaker constraints on factual precision but place strong emphasis on emotional alignment, prosodic variation, and natural conversational flow.
This shift in task demands reshapes the leaderboard: Qwen~2.5-Omni attains its highest score in Social (1155.9), surpassing its performance in Education and Medical, while FunAudioLLM (1105.9) and Kimi-Audio (1100.1) demonstrate competitive strength in Entertainment.
These patterns indicate that expressive domains amplify the importance of controllable speech generation and fine-grained paralinguistic modeling.

\textbf{Finding 3: Domain sensitivity exposes complementary strengths and unavoidable trade-offs.}  
No single model dominates all domains uniformly.
Several systems exhibit pronounced specialization: FunAudioLLM performs substantially better in Entertainment (1105.9) than in Medical (876.2), whereas Qwen~2.5-Omni peaks in Social but lags behind in Education.
Such variability suggests that the S2S model's performance is inherently domain-dependent and cannot be faithfully captured by a single aggregate score.
Effective real-world S2S systems therefore require a careful balance between semantic reliability and expressive adaptability.

\subsection{Performance under Task Difficulties}
We further analyze the results across task difficulty levels and examine the underlying model architectures based on the statistics in Table~\ref{tab:level_elo}.

\textbf{Finding 1: Industrial systems consistently outperform academic systems across all difficulty levels.}
GPT-4o-realtime (1144.2) and Doubao (1136.8) clearly lead the benchmark, whereas the strongest academic model, LLaMA-Omni, achieves only 951.4 on average due to the scale of training data, resulting in a gap of nearly 200 Elo points.

\textbf{Finding 2: The performance gap widens substantially as task difficulty increases.}
At L1, LLaMA-Omni (977.7) already performs comparably to Kimi-Audio (980.6), suggesting that basic instruction following is no longer the main bottleneck.
However, once tasks require expressive generation and full paralinguistic interaction, the difference expands sharply:
At L3, GPT-4o-realtime (1241.7) and Doubao (1148.2) exceed LLaMA-Omni (920.2) by more than 300 Elo points,
and at L4, Doubao reaches 1205.8 compared to 942.4 for LLaMA-Omni.

\textbf{Finding 3: Architectural design choices explain a large portion of the remaining variation within each group.}
Stronger backbone models improve instruction following (e.g., Qwen~2.5-Omni 1072.2 vs.\ Mini-Omni 985.8 at L1),
larger encoders enhance paralinguistic perception (LLaMA-Omni 965.2 vs.\ Mini-Omni 803.0 at L2),
while vector quantization provides no clear benefit (Kimi-Audio 949.7 vs.\ Qwen~2.5-Omni 1109.1).
For expressive speech generation, flow-matching-based modeling emerges as a critical factor
(Doubao 1205.8 vs.\ LLaMA-Omni 942.4 at L4),
whereas the choice between HiFi-GAN and BigVGAN exhibits limited influence on overall performance.

Overall, these results reveal not only a substantial and growing capability gap between industrial and academic S2S systems but also clarify how specific architectural decisions influence performance as paralinguistic demands increase. 

\section{Conclusion}
We present S2S-Arena, a benchmark for evaluating speech-to-speech (S2S) models with both instruction-following ability and paralinguistic awareness. By combining a four-level interaction protocol, a two-stage data construction pipeline, and a scalable arena-style evaluation framework, S2S-Arena enables fine-grained assessment of semantic correctness and expressive quality in S2S systems. Through over 1,000 pairwise comparisons across 10 models, we reveal significant limitations in current systems, especially under complex interaction settings, and identify a pronounced performance gap between academic and industrial models. Further analysis suggests that this gap is strongly related to differences in training data, backbone models, speech encoders, and speech decoders. S2S-Arena shifts S2S evaluation from transcript-level correctness to interaction-level human alignment, enabling the study of expressive intelligence in spoken agents.

\section*{Limitations}

Despite its strengths, S2S-Arena has several limitations. 
First, although our self-instruction pipeline enables scalable data construction, the overall dataset size remains modest compared with the diversity of real-world spoken interactions, and the use of high-quality synthetic speech may bias evaluation toward models better adapted to such distributions. 
Second, the current benchmark primarily targets utterance-level and short-range interaction, and does not yet capture long-horizon phenomena such as sustained persona consistency, long-term emotional dynamics, or discourse-level coherence. 

\section*{Ethical considerations}
In this work, except for the cases where LLM is required in the main experiments that we have disclosed, we only used LLM to polish the paper, and all references were manually verified from Google Scholar or DBLP to ensure authenticity.

The manual data was collected from members of our research group. All participants were informed about the purpose of the study and provided explicit consent for their data to be used in research and publication. All data has been anonymized to remove any personally identifiable information.

Our work may involve potential risks, including misuse of generated dialogue or speech (e.g., impersonation or misleading content) and privacy concerns related to human data. To mitigate these risks, we use anonymized data and limit our experiments to controlled research settings.

\section*{Acknowledgments}
This work was supported by National Natural Science Foundation of China (Grant No. 62271432), Program for Guangdong Introducing Innovative and Entrepreneurial Teams, Grant No. 2023ZT10X044, Major Frontier Exploration Program (Grant No. C10120250085) from the Shenzhen Medical Academy of Research and Translation (SMART),  Shenzhen Medical Research Fund (B2503005),  NSFC grant 72495131, the 1+1+1 CUHK-CUHK(SZ)-GDSTC Joint Collaboration Fund, Guangdong Provincial Key Laboratory of Mathematical Foundations for Artificial Intelligence (2023B1212010001), and the International Science and Technology Cooperation Center, Ministry of Science and Technology of China (under grant 2024YFE0203000).



\bibliography{custom}

\appendix
\section{Appendix}

\subsection{S2S models}
\label{app:s2s-models}

Table~\ref{tab:speech-models-parainfo} summarizes the representative large language model (LLM) based Speech-to-Speech (S2S) systems evaluated in this work.  
We organize these models into two groups: \textit{industrial systems} and \textit{academic models}, according to their development context and the availability of architectural details.

\subsubsection{Model Categorization}
The \textit{industrial systems} correspond to large-scale commercial or production-level models, whose internal implementations are partially disclosed or remain proprietary.  
The \textit{academic models} consist of research systems with fully documented architectures and reproducible experimental configurations.

\paragraph{Industrial Systems.}
As shown in Table~\ref{tab:speech-models-parainfo}, the industrial group includes Gemini 2.5-Pro, GPT-4o-realtime and Doubao, whose detailed architectural components are not publicly available, as well as recently released large-scale systems such as FunAudioLLM, GLM-4-Voice, Kimi-Audio, Baichuan-Omni-1.5, and Qwen2.5-Omni.

\begin{table*}[!ht]
\centering
\resizebox{\linewidth}{!}{
\begin{tabular}{llll}
\hline
\textbf{Model Name} & \textbf{Backbone} & \textbf{Encoder} & \textbf{Decoder}  \\
\hline
\multicolumn{4}{l}{\textbf{(A) Industry}} \\
\hline
Gemini 2.5-Pro & Unknown & Unknown & Unknown \\
GPT-4o-realtime & Unknown & Unknown & Unknown \\
Doubao & Unknown & Unknown & Unknown \\
FunAudioLLM~\cite{speechteam2024funaudiollm}  & Qwen-2 72B & SenseVoice & CosyVoice \\
GLM-4-Voice~\citep{zeng2024glm}  & GLM-4 9B         & Whisper-large-v3 + VQ        & FlowMatch + HiFi-GAN  \\
Kimi-Audio~\cite{ding2025kimi}   & Qwen-2.5 7B      & Whisper-large-v3 + VQ        & FlowMatch + BigVGAN               \\
Baichuan-Omni-1.5~\cite{li2025baichuan}  & Qwen-2.5 7B      & Whisper-large-v3 + RVQ        & FlowMatch + HiFi-GAN               \\ 
Qwen2.5-Omni~\citep{xu2025qwen2}    & Qwen-2.5 7B      & Whisper-large-v3 + Adaptor   & FlowMatch + BigVGAN  \\
\hline
\multicolumn{4}{l}{\textbf{(B) Academia}} \\
\hline
SpeechGPT~\citep{speechGPT}      & LLaMA 7B         & mHuBERT                       & HiFi-GAN                           \\
AnyGPT~\citep{zhan2024anygpt}    & LLaMA-2 7B       & SpeechTokenizer              & SoundStorm             \\
Freeze-Omni~\citep{wang2024freeze}  & Qwen-2 7B        & Chunk-wise Speech Encoder    & NAR + AR + TiCodec         \\
Mini-Omni~\citep{xie2024mini}       & Qwen-2 0.5B      & Whisper-small-v3 + Adaptor   & SNAC Audio Decoder                 \\
LLaMA-Omni~\citep{fang2024llama}    & LLaMA-3.1 8B     & Whisper-large-v3 + Adaptor   & NAR + HiFi-GAN  \\
\hline
\end{tabular}
}
\caption{Representative LLM-based Speech2Speech models from industry and academia. LSLM~\cite{ma2024language} and Moshi~\cite{defossez2024moshi} are excluded for fair comparison as they do not use LLM backbones.}
\label{tab:speech-models-parainfo}
\end{table*}

These systems employ diverse LLM backbones, ranging from GLM-4 and Qwen-2.5 (7B) to the 72B-scale Qwen-2 backbone used by FunAudioLLM.  
For speech encoding, most industrial models adopt Whisper-large-v3 as the acoustic front-end, optionally combined with vector quantization (VQ) or residual vector quantization (RVQ), while FunAudioLLM utilizes the proprietary SenseVoice encoder.  
On the decoding side, the dominant design integrates FlowMatch-based neural vocoders with HiFi-GAN or BigVGAN for high-fidelity waveform generation, whereas FunAudioLLM relies on the CosyVoice synthesis engine.

\paragraph{Academic Models.}
The academic group covers SpeechGPT, AnyGPT, Freeze-Omni, Mini-Omni, and LLaMA-Omni, representing a broad spectrum of research-oriented S2S designs.

SpeechGPT and AnyGPT adopt discrete speech tokenization mechanisms via mHuBERT and SpeechTokenizer, respectively, enabling direct interaction between speech representations and LLM token spaces.  
Freeze-Omni introduces a chunk-wise speech encoder with a hybrid non-autoregressive and autoregressive decoding strategy combined with TiCodec for efficient speech synthesis.  
Mini-Omni, LLaMA-Omni, and Qwen-based academic systems rely on Whisper-family encoders equipped with lightweight adaptor modules, and integrate modern neural vocoders such as SNAC and HiFi-GAN.

\subsubsection{Architectural Coverage}
Collectively, the evaluated models span a wide range of LLM backbones, speech encoders, and speech decoders, covering both discrete and continuous speech representations as well as multiple neural vocoding paradigms.  
This diverse architectural landscape provides a comprehensive evaluation testbed for assessing S2S models with respect to both semantic understanding and paralinguistic modeling capabilities.

\subsection{Self-instructed Script Augmentation Pipeline}
\label{app:audio_script}

The scripts used in S2S-Arena are automatically constructed via a self-instruct-based augmentation pipeline.
Given an initial seed set, the pipeline expands each subset into a fixed-size collection of synthetic scripts with controlled structure, task diversity, and paralinguistic levels.

\begin{tcolorbox}[aibox={Level Guide}]
Level definitions:

L1 Instruction-Only: follow the semantic instruction; paralinguistics are not evaluated.

L2 Perceive-In: infer paralinguistic information present in the input speech and adapt the reply accordingly (no requirement on output paralinguistics).

L3 Express-Out: obey an explicit request to embed designated paralinguistic traits in the output speech, while the input speech itself is neutral.

L4 Perceive \& Express: jointly understand input paralinguistics and reproduce appropriate paralinguistics in the output, mirroring real S2S interaction.

\end{tcolorbox}

\paragraph{Structured Generation.}
Each script is represented as a JSON object following a fixed schema:
\texttt{\{id, text, task, category, task\_description, language, text\_cn, level\}}.
The schema is enforced at generation time, and all outputs are produced in strict JSON format to ensure structural validity.
The \texttt{text} field is designed to serve as direct input for TTS systems.

\paragraph{Seed Sampling and Expansion.}
For each generation round, five seed samples are randomly selected as in-context examples.
Conditioned on these seeds, the model generates one new script that is semantically plausible, structurally consistent, and significantly different from the seeds.
The generation process is repeated until each subset reaches the target size of 50 samples.

\begin{tcolorbox}[aibox={System Prompt}]
You are an expert speech-to-speech (S2S) evaluation data curator. Your goal is to create synthetic test items that will be used to benchmark S2S large-language models both **with** and **without** paralinguistic information (e.g., prosody, emotion, background noise, speaker traits). Given several seed JSON examples, return new JSON that: 

• Follow the exact schema of the seeds (same keys, valid values).

• Reflect realistic, diverse content spanning different tasks, domains, languages, and paralinguistic conditions.

• Respect the dataset’s four difficulty levels L1-L4 (see definitions below).

• Contain only valid JSON (no comments or extra text).

\end{tcolorbox}

\paragraph{Controlled Generation with Reliability.}
Each generation round allows up to three retries to handle invalid outputs or parsing failures.
Intermediate results are immediately written to disk using a checkpoint mechanism, enabling safe recovery from interruptions and guaranteeing deterministic dataset growth.

\begin{tcolorbox}[aibox={User Prompt}]
\# Schema

\{schema\}

\# Level Guide

\{level\_guide\}

\# Seeds (5 examples)

\{seed\_json\}

\# Task

Generate one **new** example that:

1. Conform to the schema exactly (all required keys, same key order is appreciated). 

2. Are plausible and should significant different with the seeds. 

3. Use unique "id"s prefixed with "{id\_prefix}\_" followed by an incrementing integer. 

4. It should be considered that the generated **<text>** attribute can serve as a good input for the TTS model, thereby generating problem speech. 

5.  **<category>** should be {category}, **<tasks>** could be various and **<language>** should be Chinese or English. 

6. Return ONLY the JSON, nothing else.

\end{tcolorbox}

\paragraph{Batch Processing.}
The pipeline supports both single-file and full-corpus processing.
All subsets are independently augmented following the same procedure, allowing scalable and fully automated construction of the final audio script collection.

\subsection{Details of S2S-Arena Dataset}
~\label{app:dataset}

\subsubsection{Distribution of Samples in Seed Dataset}~\label{app:sample}
The S2S-Arena (Seed) set contains 293 manually crafted samples spanning four domains: \textit{Education}, \textit{Social Companionship}, \textit{Entertainment}, and \textit{Medical Consultation}. These samples cover a total of 19 distinct tasks, each annotated with a complexity level ranging from L1 (simplest) to L4 (most complex). The distribution of samples across tasks and levels is presented in Table~\ref{tab:sample}. We note that high-complexity tasks (L3 and L4) constitute the majority of the dataset, reflecting the practical demand for nuanced paralinguistic reasoning in speech interaction.

To provide a clearer understanding of each task, Table~\ref{tab:tasks} details the evaluation targets. For example, tasks such as \textit{Emotion Recognition and Expression} and \textit{Sarcasm Detection} aim to assess the model's ability to perceive and react to subtle emotional and prosodic cues, while others like \textit{Language Consistency} and \textit{Cross-lingual Emotional Translation} focus on multilingual and affective alignment. Tasks under the \textit{Entertainment} domain, such as \textit{Singing Capability} and \textit{Stand-up Comedy}, evaluate the generative creativity and expressive diversity of speech models.

\begin{table*}[!ht]
\centering

\begin{tabular}{llccccc}
\hline
\textbf{Domain} & \textbf{Task} & \textbf{L1} & \textbf{L2} & \textbf{L3} & \textbf{L4} & \textbf{Total} \\ \hline
\multirow{7}{*}{\textbf{Education}} & Pronunciation correction & 0 & 3 & 3 & 3 & 9 \\ 
 & Emphasis control & 0 & 2 & 3 & 1 & 6 \\
 & Rhythm control & 0 & 3 & 6 & 3 & 12 \\
 & Polyphonic word comprehension & 0 & 6 & 3 & 3 & 12 \\
 & Pause and segmentation & 0 & 2 & 2 & 2 & 6 \\ 
 & Cross-lingual emotional translation & 0 & 3 & 5 & 10 & 18 \\ 
 & Language consistency & 24 & 2 & 3 & 3 & 32 \\ \hline
\multirow{4}{*}{\textbf{Social Companionship}} & Implication understanding & 0 & 4 & 3 & 3 & 10 \\ 
 & Sarcasm detection & 0 & 3 & 3 & 2 & 8 \\
 & Identity-based response & 0 & 12 & 4 & 4 & 20 \\ 
 & Emotion recognition and expression & 0 & 4 & 4 & 27 & 35 \\ \hline
\multirow{7}{*}{\textbf{Entertainment}} & Singing capability & 0 & 3 & 5 & 2 & 10 \\
 & Natural sound simulation & 0 & 10 & 5 & 10 & 25 \\
 & Poetry recitation & 3 & 5 & 6 & 5 & 19 \\
 & Role-playing & 0 & 3 & 3 & 2 & 8 \\
 & Storytelling & 0 & 5 & 3 & 5 & 13 \\ 
 & Tongue twisters & 0 & 3 & 3 & 7 & 13 \\ 
 & Stand-up comedy/skit performance & 0 & 5 & 8 & 5 & 18 \\ \hline
\textbf{Medical Consultation} & Health consultation & 5 & 3 & 8 & 3 & 19 \\  \hline
\textbf{Total}&  & 32 & 81 & 80 & 100 & 293
\\ \hline
\end{tabular}
\caption{Distribution of Samples in S2S-Arena Seed Dataset.}\label{tab:sample}

\end{table*}

\begin{table*}[!ht]
\centering
\resizebox{\textwidth}{!}{

\begin{tabular}{llp{11cm}}
\hline
\textbf{Domain} & \textbf{Task} & \textbf{Evaluation Target} \\ \hline
\multirow{7}{*}{\textbf{Education}} & Pronunciation correction & Can the model correct inaccurate pronunciations? \\ 
& Emphasis control & Can the model understand stress emphasis and emphasize specific content with the right stress? \\
& Rhythm control & Can the model adjust the output pace, speaking faster or slower as required? \\
 & Polyphonic word comprehension & Can the model accurately understand polyphonic word? \\
 & Pause and segmentation & Can the model accurately pause and segment in ambiguous cases? \\ 
 & Cross-lingual emotional translation & Can the model accurately convey emotions during translation? \\ 
 & Language consistency & Does the model respond in the same language as the query when asked in different languages? \\ \hline
\multirow{4}{*}{\textbf{Social Companionship}} & Implication understanding & Can the model respond humorously, understanding implied meanings? \\ 
 & Sarcasm detection & Can the model detect sarcasm in phrases like “You’re amazing!”? \\
 & Identity-based response & Can the model adapt responses based on the user’s age (child, adult, elderly) and handle identity-based queries? \\ 
 & Emotion recognition and expression & Can the model recognize emotions and provide appropriate responses based on different emotions? \\ \hline
\multirow{7}{*}{\textbf{Entertainment}} & Singing capability & Can the model sing a song upon request? \\
 & Natural sound simulation & Can the model simulate certain natural sounds? \\
 & Poetry recitation & Can the model recite poems? \\
 & Role-playing & Can the model simulate a character with specific age, gender, accent, and voice tone? \\
& Storytelling & Can the model narrate a story with emotional depth? \\ 
 & Tongue twisters & Can the model correctly pronounce a given tongue twister? \\ 
 & Stand-up comedy/skit performance & Can the model perform a skit, playing both roles in a comedic dialogue? \\ \hline
\multirow{1}{*}{\textbf{Medical Consultation}} 
 & Health consultation & Can the model provide general health advice? \\  \hline
\end{tabular}
}
\caption{Task Description Across Four Domains in S2S-Arena Seed Dataset.}\label{tab:tasks}

\end{table*}

\subsubsection{Distribution of Samples in Augment Dataset}
\label{app:task-distribution}

To analyze the diversity of the newly generated Augment dataset, we perform a systematic task normalization and clustering procedure on the 950 automatically constructed samples.

Each sample is associated with a natural-language task description.  
We first apply a rule-based normalization process, including lowercasing, punctuation removal, and whitespace normalization.  
To further merge semantically identical or near-duplicate task names, we apply fuzzy matching based on string containment and Levenshtein edit distance (maximum distance = 3).  
After this normalization and merging step, the number of distinct task descriptions is reduced to 503.

We then perform unsupervised task clustering on the normalized task set.  
Each task is represented using TF--IDF features with unigram and bigram n-grams.  
We apply K-Means clustering, where the number of clusters $k$ is selected by maximizing the silhouette score within the range $k \in [3, 30]$.  
This procedure yields a final partition of the task space into 28 clusters.

\begin{figure}[ht]
    \centering
    \includegraphics[width=\linewidth]{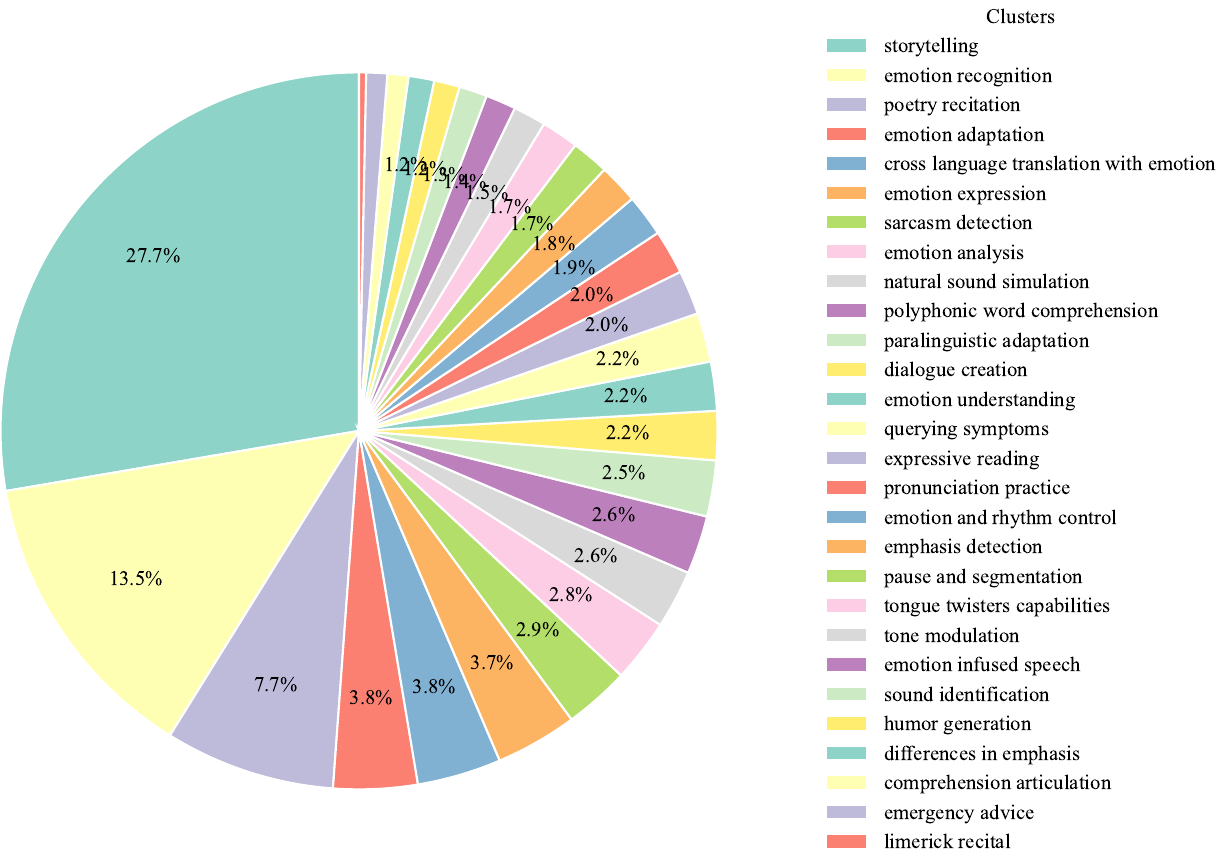}
    \caption{The task distribution in S2S-Arena augment dataset.}
    \label{fig:task-pie}
\end{figure}

Figure~\ref{fig:task-pie} summarizes the resulting task distribution over the 950 samples.  The distribution reveals that a substantial portion of the dataset focuses on paralinguistic and expressive interaction behaviors, including emotion recognition, emotion adaptation, expressive reading, tone modulation, emotion-infused speech, and other related phenomena, highlighting the dataset's emphasis on modeling rich paralinguistic information in speech interaction.

\subsection{Experimental Details}
\label{app:Experimental Details}
To ensure fair evaluation, all input samples are resampled to the appropriate sampling rate required by each speech model (e.g., 22,500 Hz for SpeechGPT and 16,000 Hz for Doubao). Model implementations are obtained from their official GitHub repositories or websites. For FunAudioLLM, GPT-4o-realtime, Doubao, and Qwen-Omni, we utilize their official APIs; all other models are executed locally using an NVIDIA A6000 GPU with 48 GB memory for inference. For models that do not natively support audio file input, we adapt the official codebases to integrate with our evaluation pipeline. All inference code will be publicly released upon acceptance of this paper.

\subsection{Preliminary Pilot Study}~\label{app:elo_rank}

We conducted a preliminary pilot study using a manually constructed Seed dataset, comprising 432 pairwise comparisons annotated by 19 Mandarin-speaking researchers (from undergraduate to postdoctoral level), all with IELTS speaking scores above 6.5. The results are presented in Table~\ref{tab:small}. In addition to the main models (GPT-4o-Realtime, SpeechGPT, Mini-Omni, and LLaMA-Omni), we also included two additional cascade models: Pipeline (4o) and FunAudioLLM (4o). Both models employ the same LLM and speech decoder (GPT-4o and CosyVoice), but differ in their speech encoders: Pipeline (4o) uses Whisper, while FunAudioLLM (4o) employs SenseVoice. These additions were intended to examine the impact of different speech encoders on model performance.

To ensure the high quality of manual judgment, we collected 10\% of samples in 432 comparisons that were independently labeled by two human annotators. The inter-annotator agreement of them achieves 83.7\%.

\paragraph{Manual Judging Findings.}

The preliminary results are largely consistent with our main evaluation findings. GPT-4o-Realtime outperforms all other open-source models, particularly excelling in domains such as education and medical consultation, which demand substantial knowledge grounding. It also ranks highly in social companionship tasks, likely due to its strong handling of paralinguistic cues. Interestingly, its performance in entertainment scenarios is relatively poor. Upon closer inspection, we observed that the model often refuses to engage in tasks it deems beyond its capabilities.

Furthermore, consistent with our main experiments, we observe that stronger speech encoders, such as Whisper compared to SenseVoice, tend to significantly boost overall system performance, as evidenced by the performance of FunAudioLLM (4o) and Pipeline (4o).

\paragraph{Human vs. Automatic Evaluation Consistency.}

To assess the reliability of our automatic evaluation framework, we first compare human judgments with two automatic evaluators: Gemini 2.5-Pro\footnote{\url{https://cloud.google.com/vertex-ai/generative-ai/docs/models/gemini/2-5-pro}} and Qwen 2.5-Omni in the Seed dataset. 

We design our automatic evaluation prompt following the Gemini API specifications, consisting of two components: \textit{Text Content Instruction} and \textit{Audio Evaluation Prompt}. For the audio prompt, the actual input to the model is a synthesized audio file in which the target audio segments are inserted at designated positions. For clarity, we include the text version of the audio prompt below. The text instruction is provided to the model as plain text.

\begin{tcolorbox}[aibox={Text Content Instruction}]
Please follow the audio instruction to generate the response.
\end{tcolorbox}

\begin{tcolorbox}[aibox={Audio Evaluation Prompt ((Text Modality))}]
I will provide an input audio and two corresponding response audios. Please evaluate which response is better. You only need to reply with 'First one wins' or 'Second one wins.'

Here is the input audio: [input audio], the first response: [output audio 1], and the second response: [output audio 2].

My requirement: A. When comparing the quality of two output audios, you need to check:

1. Check the instruction alignment: Does output audio follow the instructions of the input audio or complete the corresponding task?
    
2. Check the expressiveness: Does the output audio recognize the paralinguistic information (such as vocal tone, speaking speed, emotion, etc.) in the input audio or respond to such paralinguistic information?
    
3. Check the quality of the output audio: the sound (such as fluency, integrity, naturalness of speech).
\end{tcolorbox}

As shown in Table~\ref{tab:auto_eval_agreement}, Gemini 2.5-Pro achieves a Cohen's Kappa~\cite{cohen1960kappa} of 0.6553 and an observed agreement rate of 82.87\%, indicating substantial alignment with human preferences. In contrast, Qwen2.5-Omni achieves a moderate agreement with a Kappa of 0.4667 and 73.15\% accuracy. We thus adopt Gemini 2.5-Pro as the default automatic evaluator in subsequent experiments that closely cooperate with the intra-human annotation agreement.

\begin{table}[!ht]
\centering
\resizebox{\linewidth}{!}{
\begin{tabular}{lccccc}
\hline
 Model          & Overall & Edu. & Social & Enter. & Med. \\ \hline
 GPT-4o-Realtime         & \textbf{1365}    & \textbf{1185}      & 1064                 & 970           & \textbf{1146}                 \\ 
Pipeline (4o)           & 1207    & 1065      & 995                  & 1069          & 1077                 \\ 
FunAudioLLM (4o)      & 1025    & 1105      & \textbf{1077}        & 850           & 993                  \\
SpeechGPT         & 849     & 906       & 919                  & \textbf{1095} & 929                  \\
Mini-Omni                & 841     & 857       & 1000                 & 1041          & 943                  \\ 
LLaMA-Omni               & 714     & 882       & 945                  & 975           & 911                  \\ \hline
\end{tabular}
}
\caption{Elo Rankings from Small-Scale Human Evaluation on S2S-Arena (Seed).}
\label{tab:small}
\end{table}

\end{document}